\documentclass[runningheads]{llncs}

\usepackage{eccv}

\usepackage{eccvabbrv}

\usepackage[pdftex]{graphicx}
\usepackage{subcaption}
\usepackage{float}
\usepackage[justification=raggedright]{caption}	
\usepackage{lscape}                                         
\usepackage{wrapfig}

\usepackage[lined,ruled,linesnumbered]{algorithm2e}

\usepackage{booktabs}                   
\usepackage{multirow}

\usepackage{makecell}

\usepackage{paralist}
\usepackage{enumitem}
\setlist[enumerate]{itemsep=0mm}

\usepackage{bm}                          
\usepackage{epsfig}                      
\usepackage{mathtools}
\usepackage{amssymb,amsmath}   

\usepackage{units}
\usepackage{color}

\usepackage{comment}

\usepackage[hyphens]{url}
\usepackage[breaklinks,colorlinks,linkcolor=blue,citecolor=blue,urlcolor=blue,bookmarks=false]{hyperref}
\usepackage{xspace}
\usepackage{setspace}

\usepackage{gensymb}

\usepackage{soul}
\usepackage{svg}

\DeclareMathAlphabet{\altmathcal}{OMS}{cmsy}{m}{n}
\DeclareMathAlphabet{\mathbfit}{OT1}{ptm}{bx}{it}

\newlength\paramargin
\newlength\figmargin

\newlength\secmargin
\newlength\figcapmargin
\newlength\tabcapmargin

\newcommand{\topic}[1]
{
\vspace{1mm}\noindent\textbf{#1}
}

\long\def\ignorethis#1{}

\newbox\jsavebox%

\makeatletter
\newcommand{\providelength}[1]{%
  \@ifundefined{\expandafter\@gobble\string#1}
   {
    \typeout{\string\providelength: making new length \string#1}%
    \newlength{#1}%
   }
   {
    \sdaau@checkforlength{#1}%
   }%
}

\newcommand{\sdaau@checkforlength}[1]{%
  \edef\sdaau@temp{\expandafter\sdaau@getfive\meaning#1TTTTT$}%
  \ifx\sdaau@temp\sdaau@skipstring
    \typeout{\string\providelength: \string#1 already a length}%
  \else
    \@latex@error
      {\string#1 illegal in \string\providelength}
      {\string#1 is defined, but not with \string\newlength}%
  \fi
}
\def\sdaau@getfive#1#2#3#4#5#6${#1#2#3#4#5}
\edef\sdaau@skipstring{\string\skip}
\makeatother

\usepackage[capitalize]{cleveref}
\crefname{section}{Sec.}{Secs.}
\Crefname{section}{Section}{Sections}
\Crefname{table}{Table}{Tables}
\crefname{table}{Tab.}{Tabs.}

\def\xi{\mathbf{x}_i}

\graphicspath{{figures}, {examples}}

\usepackage{hyperref}

\newcommand{\pare}[1]{\left(#1\right)}

\begin{document}

\title{Flash-Splat: 3D Reflection Removal with \\ Flash Cues and Gaussian Splats} 

\titlerunning{Flash-Splat: 3D Reflection Removal with
Flash Cues and Gaussian Splats}


\author{Mingyang Xie$^{1}$\thanks{Equal Contribution.}\quad Haoming Cai$^{1}$$^{\star}$\quad Sachin Shah$^{1}$ \quad Yiran Xu$^{1}$\\Brandon Y. Feng$^{2}$ \quad Jia-Bin Huang$^{1}$ \quad Christopher A. Metzler$^{1}$}


\authorrunning{M.~Xie et al.}

\institute{$^{1}$University of Maryland $^{2}$Massachusetts Institute of Technology}
\institute{$^{1}$University of Maryland $^{2}$Massachusetts Institute of Technology 
\\ {\href{https://flash-splat.github.io/}{\textcolor{magenta}{https://flash-splat.github.io/}}}}




\maketitle


\begin{abstract}

We introduce a simple yet effective approach for separating transmitted and reflected light. Our key insight is that the powerful novel view synthesis capabilities provided by modern inverse rendering methods (e.g.,~3D Gaussian splatting) allow one to perform flash/no-flash reflection separation using {\em unpaired measurements}---this relaxation dramatically simplifies image acquisition over conventional paired flash/no-flash reflection separation methods. Through extensive real-world experiments, we demonstrate our method, Flash-Splat, accurately reconstructs both transmitted and reflected scenes in 3D. Our method outperforms existing 3D reflection separation methods, which do not leverage illumination control, by a large margin. 
This paper appears at {{\textcolor{magenta}{ECCV 2024}}}.



\end{abstract}

\section{Introduction}
\label{sec:intro}








We are often surrounded by scenes with transparent surfaces, most notably glass, which introduce specular reflections. 
When viewing such scenes, we see a superimposition of transmitted and reflected light. This work focuses on the unsupervised separation of a transmitted 3D scene and a reflected 3D scene.

Reflection removal and separation have received considerable attention in the computational photography community. In addition to enhancing image quality and appeal, effective reflection separation methods can improve the robustness of downstream computer vision systems used in various applications, including robot navigation, classification, and 3D surface reconstruction. 
Separating transmitted and reflected 3D scenes is vital for various virtual reality tasks, such as 3D object extraction or editing.

\begin{figure}[t!]
  \centering
   \includegraphics[width=\columnwidth]{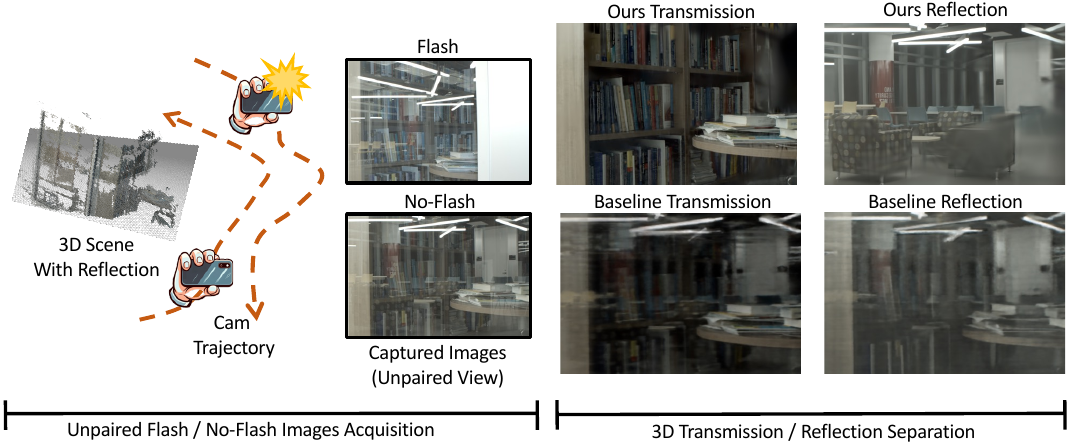}
   \caption{\textbf{Left}: We separate the 3D transmitted and reflected scenes by capturing some views with camera flash and some views with no flash. \textbf{Right}: Our proposed Flash-Splat method achieves much better separation than the state-of-the-art unsupervised 3D separation method NeRFReN~\cite{guo2022nerfren}. }
   \label{fig:teaser}
\end{figure}

Unfortunately, separating transmitted and reflected light from the sum of their intensities is a highly under-determined problem. 
To address this challenge, prior works have relied on various assumptions to perform single-image reflection removal.
For instance, they have assumed the reflection is out-of-focus~\cite{arvanitopoulos2017single,yang2019fast} or there is a noticeable double reflection caused by two sides of the glass~\cite{shih2015reflection}. However, these assumptions are not always true in real life.
Other works have leveraged videos or multi-view images for reflection removal~\cite{gandelsman2019double,hong20232,hong2021panoramic,liu2020learning,alayrac2019visual,xue2015computational,guo2014robust}.
Their advantages over single-image methods are (1) they can get ``lucky'' where some views have weaker reflections than others, and (2) they can utilize multi-view consistency to regularize the separation. However, these methods still struggle to overcome the fundamental ill-posed problem, especially under strong reflection. For example, in Figure~\ref{fig:teaser} the state-of-the-art unsupervised 3D reflection separation method, NeRFReN~\cite{guo2022nerfren}, fails to separate reflected and transmitted light from a collection of images captured under similar illumination conditions.

Introducing illumination control, i.e.,~flash/no-flash photography~\cite{lei2021robust,Xia2020FlashNormal,Xia2020FnF}, can make the reflection separation problem significantly easier. 
Intuitively, the camera flash increases the intensities of the transmitted scene while leaving the reflected scene largely intact. 
Therefore, we can recover a reflection-free transmission scene by comparing images captured with and without flash. 
The core limitation 
is that it requires paired (tightly-aligned) flash/no-flash image captures---the camera cannot move between the captures. 
This paired measurement requirement represents a major barrier to effective in-the-wild reflection separation.

In this paper, we perform flash-based reflection separation without paired measurements by leveraging the powerful novel view synthesis capabilities of recently developed inverse differentiable rendering methods. 
Specifically, during acquisition, a user captures roughly half of the views with flash on and the other half with flash off. Then, by extending the powerful Gaussian Splatting~\cite{kerbl20233dgs} technique, {we can construct 2D ``pseudo-paired'' flash/no-flash images, where one image in the pseudo flash/no-flash pair is captured, and the other one is synthesized with our 3D inverse rendering framework; we can also construct a 3D ``pseudo-pair'' of flash/no-flash 3D representations, where one 3D representation is reconstructed from only the flash images, and the other is reconstructed from only the no-flash images. The difference between the 2D pseudo-pair and the difference between the 3D pseudo-pair both serve as strong priors for the transmitted 3D scene, which significantly reduce the ill-posedness of the separation problem.} As a byproduct of our 3D inverse differentiable rendering framework,  our method, Flash-Splat, is also capable of performing novel view synthesis and depth estimation for each transmitted and reflected scene. We validate our proposed approach in real-world experiments and demonstrate its state-of-the-art performance.

Our contributions are:
\begin{itemize}
    \item We propose a robust strategy, Flash-Splat, for 3D transmission-reflection separation and scene reconstruction, using flash illumination as a physical cue without requiring paired flash/no-flash captures.
    \item We introduce novel modifications to make 3D Gaussian Splatting illumination-aware, enhancing the quality of each separated 3D scene.
    \item We show that Flash-Splat excels in separating reflection and transmission, even when baseline methods fail, over real-world scenes.
    \item We demonstrate that Flash-Splat can perform high-quality novel view synthesis and depth estimation for both the transmitted and reflected 3D scenes. 
\end{itemize}

\section{Related work}
\label{sec:related}


\vspace{-8pt}

\topic{Reflection removal.} 
Existing reflection removal methods can generally be divided into three categories: single-frame, multi-frame and polarization-based. 
Single-frame approaches~\cite{levin2002learning,levin2004separating,levin2007user,li2013exploiting,shih2015reflection,hariharan2015hypercolumns,wan2018region,yang2019fast,arvanitopoulos2017single,fan2017generic,yang2018seeing,zhang2018single,li2020single,wan2018crrn,wan2021face,wen2019single,wei2019single,kim2019single,zou2020deep,zheng2021single,dong2021location,hu2023single,li2023two,hu2021trash,wan2020reflection,Kee2024RawReflection} only take a single image and remove the reflection.
Multi-frame approaches~\cite{farid1999separating,gandelsman2019double,hong20232,hong2021panoramic,liu2020learning,alayrac2019visual,xue2015computational,guo2014robust,li2013exploiting,chugunov2024nsf} use multiple input frames as cue and produce multi-view consistent results.
Polarization-based approaches~\cite{lei2020polarized,nayar1997separation,kong2011high,kong2013physically,lyu2019reflection,li2020reflection} leverage the fact that the transmission is unpolarized while the reflection component varies when rotating the polarization filter. However, none of those methods aim to recover a 3D representation of the transmitted or the reflected scene.

\topic{3D neural scene representations.} 
To get more accurate 3D reconstruction for decomposition, we consider differentiable 3D neural representations.
Neural Radiance Fields (NeRFs)~\cite{mildenhall2020nerf,barron2022mip,chen2022tensorf,fridovich2023k} has received vast attentions in the past few years, for their accurate and consistent novel view synthesis results.
Another line of works focuses on accurate 3D geometry, so they considers Signed Distance Function (SDF)~\cite{wang2023neus2,wang2021neus} for better surface accuracy.
Recently, 3D Gaussian Splatting (3DGS)~\cite{kerbl20233dgs,zhu2023fsgs} emerges for its fast training and inference speed. 

\topic{Reflection removal by inverse rendering.}
Previous methods consider solving reflection removal using inverse 3D rendering.
ReflectionsIBR~\cite{sinha2012image} as a pioneer proposes to separate each frame into a transmission and reflection layer combined with a binary reflection mask, and tries to reconstruct the scene using an image-based rendering. 
Recently, NeRFReN~\cite{guo2022nerfren} uses a NeRF to achieve better 3D reconstruction accuracy.
NeuS-HSR~\cite{qiu2023looking}, instead of focusing on reflection separation, uses Signed Distance Functions to achieve better surface reconstruction quality. Distinct from these 3D methods, our proposed method dramatically extends these approaches by incorporating variable illumination.













\section{Method}
\label{sec:method}

\begin{figure*}[!t]
   \includegraphics[width=\textwidth]{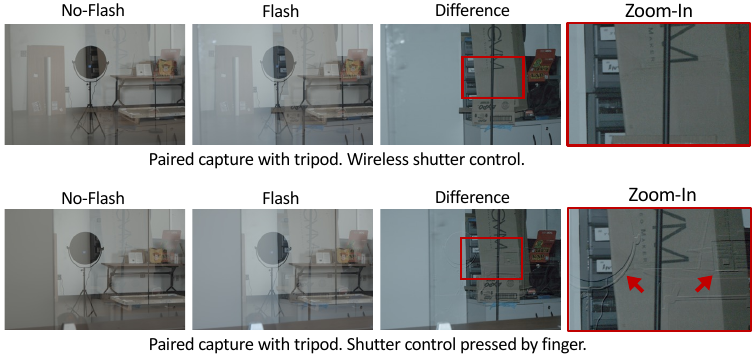}
   \caption{\textbf{Flash/No-Flash For Reflection Removal}. The difference between paired flash and no-flash images is equivalent to taking a photo with flash in a dark environment, which gives us a reflection-free image (top). 
   This is because flash increases the transmission brightness, but not the reflection brightness. 
   Notice pairs must be tightly aligned for this method to work. Even tiny vibrations such as pressing the shutter button even when using a tripod produce artifacts (bottom). 
}
   \label{fig:motivation}
\end{figure*}

\subsection{Paired 2D Flash/No-Flash}

Capturing a pair of flash and no-flash images of a scene from the same camera viewpoint allows one to reconstruct a reflection-free image. 

Reflections exist because ambient light illuminates objects in the reflected scene and reflects off the glass onto the camera sensor. The captured composite scene with no flash $\mathbf{I}_{N}$ can be modeled as 
\begin{align}
    \mathbf{I}_{N} &=  \mathbf{T}_{N}  + \beta\circ \mathbf{R}
    \label{eqn:noflash}
\end{align}
where $\mathbf{T}_{N}$ is the transmission scene with no flash, $\mathbf{R}$ is the reflection scene and $\beta$ is the reflective fraction factor (in the extreme case where the reflection is caused by a mirror, then $\beta$ could be interpreted as the mask of the mirror in the scene).  Now consider the case where the scene is captured with a flash co-located on the camera. If we assume that the scene behind glass is diffuse and that the camera flash is uniform, the camera flash will increase the intensity of all pixels proportionally. Therefore, we can formulate the flash image $\mathbf{I}_{F}$ as following:
\begin{align}
    \mathbf{I}_{F} &=  (1+ \alpha)\mathbf{T}_{N} + (\beta + {\beta}_{F})\circ \mathbf{R},
\end{align}
where $\alpha$ and ${\beta}_{F}$ represent the intensity increase of the transmitted and reflected scene due to flash, respectively. Assuming the direct reflection of the flash is outside the camera's field of view (i.e., the specular surface is not orthogonal to the camera view), the flash will have little effect on the brightness of the reflected scene. 
In common cases like glass, the impact of secondary reflections is also usually very low.
Therefore, we may approximate ${\beta}_{F}$ as close to zero, 
\begin{align}
    \mathbf{I}_{F} &\approx  (1+ \alpha)\mathbf{T}_{N} + \beta\circ \mathbf{R}.
    \label{eqn:flash}
\end{align}
As such, one technique used among photographers is subtracting a no-flash image $\mathbf{I}_{N}$~\cite{lei2021robust} from a flash image $\mathbf{I}_{F}$,
\begin{align}
    \mathbf{I}_{F} - \mathbf{I}_{N} &\approx \alpha\mathbf{T}_{N}.
    \label{eqn:reflection_free}
\end{align}
The difference is effectively a reflection-free transmitted scene scaled by some constant. Figure \ref{fig:motivation} demonstrates the impressive performance of this simple method.

Unfortunately, this process only works when we capture \textbf{paired} flash and no-flash images at the {\em same} location and orientation. Any small movement between the image pair causes the approach to break down. As illustrated in the bottom row of {Figure \ref{fig:motivation}}, even with a tripod, the slight motion caused by touching the exposure button (as opposed to using remote triggering) can introduce significant errors in the conventional flash/no-flash reflection separation process. 

\subsection{Unpaired 3D Flash/No-Flash}
\label{subsec: 3.2}
{In this work, we extend the flash/no-flash idea to 3D and thus remove the requirement of capturing paired images, which makes flash-based reflection removal significantly easier and more practical. }
{Instead of directly capturing paired multi-view images of a scene, we propose to first capture an arbitrary sequence of multi-view flash images of the scene, and then capture another sequence of multi-view no-flash images of the scene. These two sequences should be captured such that they approximately cover a similar range of perspectives.}
 

Our 3D Flash/No-flash formulation is defined as follows. Following previous notations, we consider four 3D representations in total: transmission with flash $\mathbf{T}_{F}$, transmission without flash $\mathbf{T}_{N}$, reflection $\mathbf{R}$, and the reflective fraction factor $\beta$. 
To render a target pixel in a captured image, we blend the overlapping regions of the transmitted and reflected scenes.
We then have,
\begin{align}\label{eqn:ours}
\begin{split}
    \mathbf{I}_{N} &=  \mathbf{T}_{N} + \beta \circ \mathbf{R} \\
    \mathbf{I}_{F} &=  \mathbf{T}_{F} + \beta \circ \mathbf{R} \,
\end{split}
\end{align}
for flash ($F$) images and no-flash ($N$) images. 
Even though we are only capturing unpaired flash/no-flash views now, we can still associate them by creating 2 types of ``pseudo-pairs'' to aid reflection separation.

\begin{figure}[t!]
   \includegraphics[width=\columnwidth]{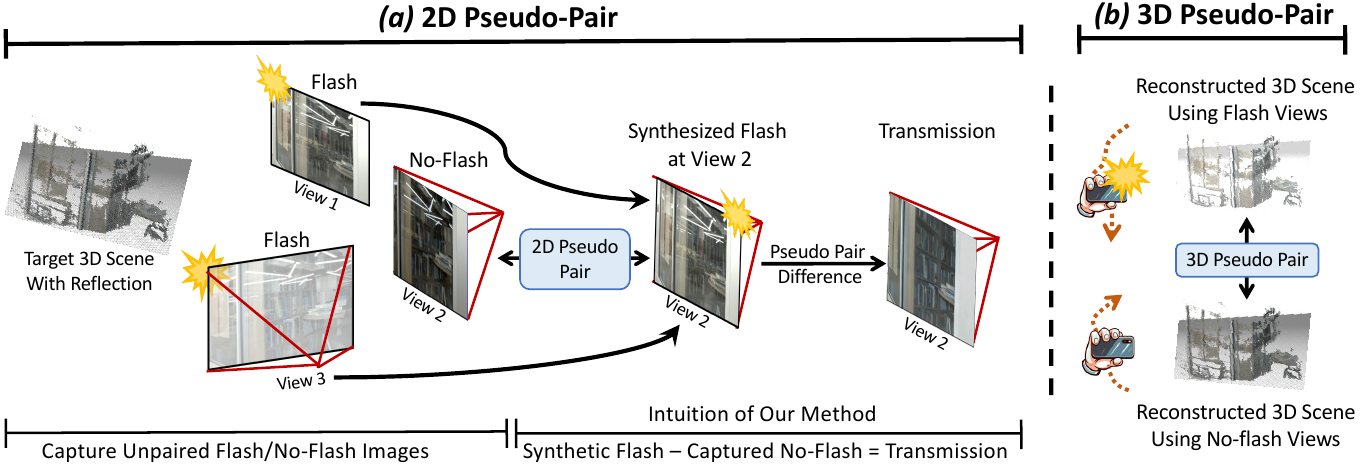}
   \caption{\textbf{Our Intuition: Construct 2D and 3D ``pseudo-pairs'' as Cues for Reflection Removal.} 
   Flash-Splat does not require \textit{paired} flash/no-flash data. During the data capture stage, we collect \textit{unpaired} flash/no-flash images from different views. In (\textcolor{blue}{a}), if we captured a no-flash image at View 2, we can learn a 3D representation of the captured flash images at other views, and then synthesize a novel view of the flash image at View 2. As such, we have created a \textbf{2D pseudo-pair} of flash and no-flash images at View 2. If we then take the difference between the pseudo-pair as in Figure~\ref{fig:motivation}, we get the transmission component of View 2 that is free of reflection. In (\textcolor{blue}{b}), we reconstruct a 3D scene with flash using only the flash images (top); we also reconstruct a 3D scene without flash using only the no-flash images (bottom). As such, we have created a \textbf{3D pseudo-pair} of flash/no-flash scenes. 
   }
   \label{fig:overview}
\end{figure}

Firstly, we can construct \textbf{2D ``pseudo-pairs''} via novel view synthesis of the missing flash/no-flash counterpart, as shown in Figure~\ref{fig:overview}\textcolor{blue}{a}. Consider a specific view where only the flash image is taken. Utilizing inverse rendering techniques, we are able to synthesize a no-flash image at this exact same view by using the no-flash images taken at neighboring views. This synthesized no-flash image and the captured flash image form a 2D pseudo-pair. The difference image between this 2D pseudo-pair should be reflection-free, just like the difference between the 2D paired flash/no-flash images, as indicated in Equation~(\ref{eqn:reflection_free}). 

Secondly, we can construct a \textbf{3D ``pseudo-pair''} by elevating the problem to the 3D space, as shown in Figure~\ref{fig:overview}\textcolor{blue}{b}. More specifically, we can reconstruct a 3D scene with flash and another without flash, using only the views captured with flash and only the views captured without flash, respectively. We name these two reconstructed scenes as ${\mathbf{I}_{F}^{{Rec}}}$ and ${\mathbf{I}_{N}^{{Rec}}}$, to differentiate them with the ground truth 3D flash/no-flash scenes $\mathbf{I}_{F}$ and $\mathbf{I}_{N}$. ${\mathbf{I}_{F}^{{Rec}}}$ and ${\mathbf{I}_{N}^{{Rec}}}$ form a 3D pseudo-pair, as they are the same scene except that the transmitted part of ${\mathbf{I}_{F}^{{Rec}}}$ is brighter due to the flash. A 3D pseudo-pair difference can be used as a cue for the transmitted scene. Nevertheless, as ${\mathbf{I}_{F}^{{Rec}}}$ and ${\mathbf{I}_{N}^{{Rec}}}$ are separately reconstructed from 2 unpaired sets of data, they will be misaligned, thus the word ``pseudo''.  

As such, we obtain important ``flash cues'' from the 2D and 3D pseudo-pairs, and use them as the high-level intuitions for our proposed method.

\subsection{Proposed Pipeline for 3D Reflection Separation}
\label{subsec: 3.3}
In this subsection, we first explain how to incorporate flash cues to guide our reconstruction, then describe our overall optimization framework, and finally discuss how to adapt the loss functions to accommodate the RAW input images.

\topic{Regularizing Reflection Separation using Flash Cues.} 
\label{subsubsec: linear_2d}
While our high-level intuition is to construct pseudo-pairs as cues for reflection-free images, in this work, we want to reconstruct both the transmitted scene and the reflected scene. Therefore, we do not explicitly calculate the difference between the pseudo-pairs, but rather, use it as a regularization term to guide separation optimization.
As shown in Equation~\eqref{eqn:flash}, $\mathbf{T}_{F}$ is expressed as $\mathbf{T}_{N}$ multiplied by a scalar $(1+\alpha)$, which enforces a linear relationship between them. Therefore, we choose to enforce the linearity between $\mathbf{T}_{F}$ and $\mathbf{T}_{N}$, which is equivalent to enforcing the constraint that the flash/no-flash difference is reflection-free.

While in the ideal case, $\mathbf{T}_{F}^{}$ and $\mathbf{T}_{N}$ should form a strictly linear relationship, in reality, the camera flash might not be perfectly uniform; there is also a chance the secondary reflection of the flash does hit the camera sensor. As a result, this relationship between $\mathbf{T}_{F}$ and $\mathbf{T}_{N}$ should be close to linear, but might not perfectly linear. Therefore, we chose not to use this hard constraint, but use the Pearson Coefficient, which measures the linearity between  $\mathbf{T}_{F}$ and $\mathbf{T}_{N}$:
\begin{equation}\label{eqn:linearity}
    \mathcal{L}_{linearity} = -\frac{\text{cov}\pare{\mathbf{T}_{N},\mathbf{T}_{F}}}{\sqrt{\text{var}\pare{\mathbf{T}_{N}}\text{var}\pare{\mathbf{T}_{F}}}}  
\end{equation}
By minimizing this loss term, we encourage the 3D Gaussians to learn a reflection-free transmission, therefore reducing the ill-posedness of the separation.

{Notably, while the analysis above holds true for both the 3D pseudo-pair (see Figure~\ref{fig:overview}\textcolor{blue}{b}) and the 2D pseudo-pairs (see Figure~\ref{fig:overview}\textcolor{blue}{a}), we only apply this linearity regularization to the 2D pseudo-pairs, as it is more straightforward to measure the linearity of images than 3D representations.}

\topic{Initializing 3D Representations Using Flash Cues.}
\label{subsubsec: init_3d}
Now we show how to utilize the 3D pseudo-pair to aid reflection separation. As illustrated in Figure~\ref{fig:motivation}\textcolor{blue}{b}, the 3D pseudo-pair, namely ${\mathbf{I}_{F}^{{Rec}}}$ and ${\mathbf{I}_{N}^{{Rec}}}$, are 3D representations of the target scene reconstructed from the flash views and no-flash views, respectively. Their difference should be the reflection-free 3D transmitted scene. However, given the highly ill-posed nature of the 3D scene reconstruction problem, it is very likely that the contents in ${\mathbf{I}_{F}^{{Rec}}}$ and ${\mathbf{I}_{N}^{{Rec}}}$ do not correspond with each other. As such, this difference between ${\mathbf{I}_{F}^{{Rec}}}$ and ${\mathbf{I}_{N}^{{Rec}}}$ should be viewed as a very rough estimate of the transmitted scene. Therefore, we decide to only use it to initialize the 3D representations $\mathbf{T}_F$, $\mathbf{T}_{N}$, $\mathbf{R}$, and $\beta$ for better convergence. 

We use 3DGS \cite{kerbl20233dgs} as the 3D representation architecture, which is normally initialized from sparse point clouds. We first use structure from motion, e.g.,~\cite{schonberger2016pixelwise}, to obtain the sparse point clouds of the 3D pseudo-pair ${\mathbf{I}_{F}^{{Rec}}}$ and ${\mathbf{I}_{N}^{{Rec}}}$. Then we roughly align them via linear transformation to compensate for the difference in the camera coordinate systems. Afterwards, we compare these two sets of point clouds: for points in regions with increased intensities, we classify them as ``transmitted points''; for points in regions with unchanged intensities, we classify them as ``reflected points''. Finally, we initialize the 3DGSs for $\mathbf{T}_F$, $\mathbf{T}_{N}$, and $\beta$ from the ``transmitted points'', and the 3DGS for $\mathbf{R}$ from the ``reflected points''.

Note that this way of initialization relies on the 3D representation's compatibility with point clouds. It does not work with implicit neural 3D representations like NeRF~\cite{mildenhall2020nerf}. When using NeRF as our 3D representations, we just randomly initialize the neural network and only rely on the previously discussed linearity regularization using 2D pseudo-pairs, which would still achieve better reflection removal performance than baselines, as will be shown in Section~\ref{sec:ablations}.

\begin{figure}[t!]
   \includegraphics[width=\columnwidth]{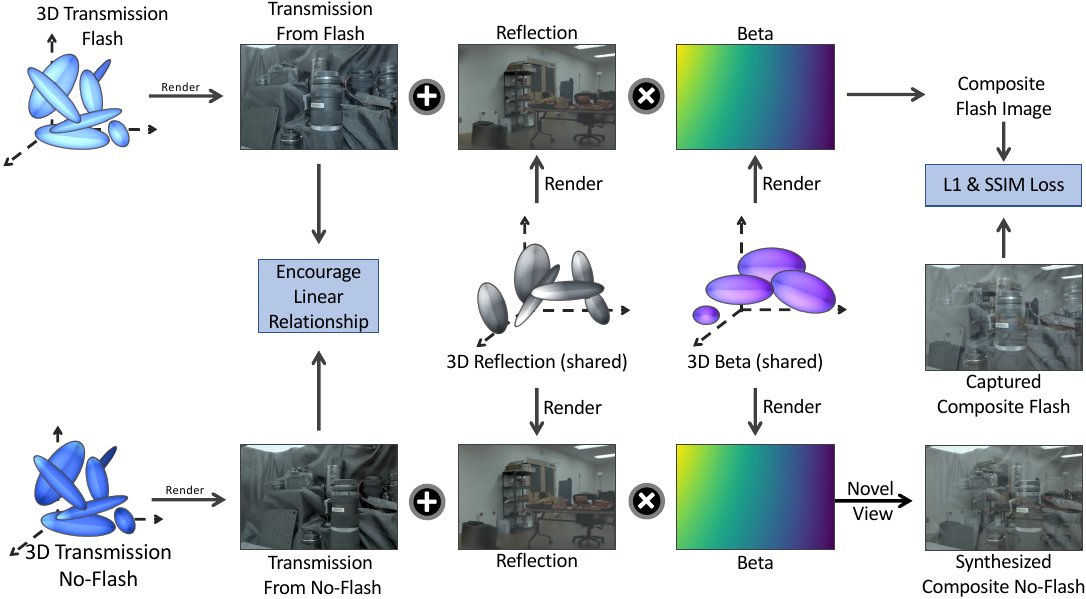}
   \caption{\textbf{Method Overview.} 
   We use 4 3DGSs~\cite{kerbl20233dgs} as our 3D representations for the transmitted scene with flash $\mathbf{T}_{F}$, the transmitted scene with no flash $\mathbf{T}_{N}$, the reflected scene $\mathbf{R}$ and the reflective fraction map $\beta$. Based on the Flash/No-flash technique, $\mathbf{R}$ and  $\beta$  are shared between the flash image and the no-flash image, while the relationship of $\mathbf{T}_{F}$ and  $\mathbf{T}_{N}$ is close to linear. We initialize these 4 3DGSs using cues from the 3D pseudo-pair (see Figure~\ref{fig:overview}\textcolor{blue}{b} and Section~\ref{subsubsec: init_3d}). In each iteration of optimization, our method operates on a single view.
   {This figure, for instance, shows a view where we captured a flash image. There is NO no-flash image captured at this view.}
   As shown in the top row, we use  $\mathbf{T}_{F}$,  $\mathbf{R}$, and $\beta$ to render a flash image of this particular view and calculate losses with the captured ground truth flash image. Additionally, based on the cues from 2D pseudo-pairs, we calculate the Pearson linearity loss between  $\mathbf{T}_{F}$ and $\mathbf{T}_{N}$ to encourage the linearity between them (see Figure~\ref{fig:overview}\textcolor{blue}{a} and Section~\ref{subsubsec: linear_2d}). We then back-propagate the gradients and update the weights of the 4 3DGSs.}
   \label{fig:pipeline}
\end{figure}

\topic{Overall Optimization Framework.} As shown in Figure~\ref{fig:pipeline}, in each iteration of optimization, Flash-Splat operates on a single view. 
{If we captured a flash image at this view (meaning that NO no-flash image was captured at this view),
we follow Equation~\eqref{eqn:ours} and use our 3D representations $\mathbf{T}_{F}$,  $\mathbf{R}$, and $\beta$ to render a flash image at this same view. Then we calculate the loss between the rendered flash image and the captured ground truth flash image (more on this in the next paragraph). Additionally, we also calculate the Pearson linearity loss between images rendered from $\mathbf{T}_{F}$ and $\mathbf{T}_{N}$ at this view (the 2D pseudo-pair).} We then back-propagate the gradients and update the weights of the $4$ 3D representations $\mathbf{T}_{F}$, $\mathbf{T}_{N}$, $\mathbf{R}$, and  $\beta$. In the next iteration, we perform similar computations with flash and no-flash swapped. By doing such alternative optimization, we are using the loss with the captured ground truth images to supervise the novel view synthesis ability of our 3D representations, while using the Pearson linearity loss to implicitly enforce the flash/no-flash prior.

\topic{Gamma Corrected Loss Function.}
\label{sec:loss}
Our complete loss function is:
\begin{equation} \label{eqn:total_loss}
    \mathcal{L} = \lambda_1 \mathcal{L}_1 + \lambda_2 \mathcal{L}_{DSSIM} + \lambda_3 \mathcal{L}_{linearity} + \lambda_4 \mathcal{L}_{depth} \,,
\end{equation}
where  $\mathcal{L}_1$ and DSSIM~\cite{Baker:DSSIM} are computed between the captured RGB images and the rendered images; $\mathcal{L}_{depth}$ is a depth smoothness regularization adopted from NeRFReN~\cite{guo2022nerfren}; $\lambda_{1-4}$ are weightings for these 4 loss terms, respectively. 

It is imperative to understand that for the difference of the flash/no-flash discrepancy to work, Equations~\ref{eqn:noflash} through \ref{eqn:linearity} must be operational within the RAW image space. Nevertheless, our objective is for the $\mathcal{L}_1$ loss and DSSIM loss to be applied to tone-mapped images, as~\cite{mildenhall2022nerfdark} demonstrates that learning within the RAW space predisposes the model to bias in favor of brighter pixels while neglecting the darker ones. Furthermore, it is our intention for the 3DGS model to output tone-mapped images, a decision driven by empirical observations indicating a potential underperformance when learning is conducted in the RAW domain. Consequently, we modify our loss function to incorporate $\mathcal{L}_1$ and DSSIM calculations as follows:
\begin{align}\label{eqn:raw_loss}
    &\mathcal{L} = \mathcal{L}\pare{\gamma\pare{\mathbf{I}^{raw}}, \gamma\pare{\gamma^{-1}\pare{\mathbf{T}} + \beta \gamma^{-1}\pare{\mathbf{R}}} }
\end{align}
where $\mathcal{L}$ can be either $\mathcal{L}_1$ or DSSIM loss, and we chose $\gamma(\mathbf{x})=\mathbf{x}^{0.22}$ as gamma correction to tone-map RAW images.

\section{Experimental Details}
\label{sec:experimental_Details}




\subsection{Dataset Collection}
We captured a flash/no-flash dataset using a Canon Rebel R7 camera. Our camera settings included a fixed 0.25-second exposure time, 200 ISO sensitivity, an f-number of 5.6, and fixed white balancing. For each scene, we collected $30$ images equally split into two categories: $15$ with the built-in flash and $15$ without. The typical distance between a flash view and the closest no-flash view is $5-10$ centimeters, with $1-6$ meters separating the camera from the transmission scene objects.
For a fair comparison with the baselines, we captured paired flash/no-flash data for a few scenes using a tripod, such that our method and the baselines use exactly the same views, with the only data difference being our method uses half flash and half no-flash views. Note that our method does not see any paired flash/no-flash views.
We process the raw images through a standard image signal processor consisting of white balancing, tone-mapping and gamma correction. 
We run COLMAP~\cite{schonberger2016pixelwise} on each method's corresponding input images
to obtain camera poses and sparse point cloud estimations.
\label{sec:setup}

\subsection{Baselines}

We choose 2 baselines: NeRFReN~\cite{guo2022nerfren}, an unsupervised multi-view method based on 3D inverse rendering, and Dong et al.~\cite{dong2021location}, a supervised deep learning approach.
For fair comparison, we run both baselines twice, once on all flash images, and once on all no-flash images. This is to show that our method does not perform better because we use flash, but rather, because we use flash/no-flash cues.

\begin{figure}[t!]
   \includegraphics[width=\columnwidth]{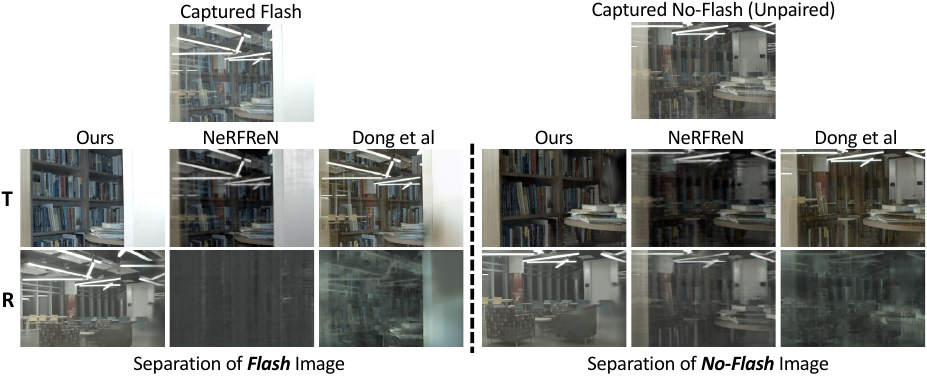}
   \caption{\textbf{The Office scene.}  Top, middle, and bottom rows are the captured images, separated transmissions, and separated reflections, respectively. Our reflection separation approach is far more effective than NeRFReN~\cite{guo2022nerfren} and Dong et al~\cite{dong2021location}.}
   \label{fig:compare_baselines_office}
\end{figure}

\begin{figure}[t!]
   \includegraphics[width=\columnwidth]{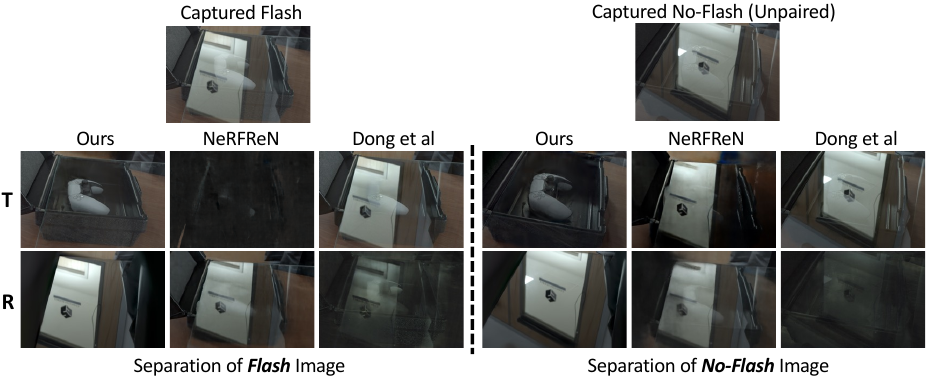}
   \caption{\textbf{The Game Controller Scene.} Top, middle, and bottom rows are the captured images, separated transmissions, and separated reflections, respectively. Our reflection separation approach is far more effective.}
   \label{fig:compare_baselines_controller}
\end{figure}

\subsection{Architecture and Optimization details.} 
\label{subsec:4.3_arch}
Flash-Splat was implemented in PyTorch and run on an NVIDIA A6000 GPU.
{ Our $4$ 3DGSs ($\mathbf{T}_F$, $\mathbf{T}_{N}$, $\mathbf{R}$, and $\beta$) have no shared parameters and are optimized with the same hyperparameters. Our implementation of 3DGS follows FSGS~\cite{zhu2023fsgs}, a variant of 3DGS that supports feedforward settings (the original 3DGS~\cite{kerbl20233dgs} is not intended for feedforward scenes). Each 3DGS is initialized with $350$K Gaussians and grow up to $500$K. We optimize all 3DGSs for $5000$ iterations on our flash/no-flash images sized $1200\times 800$ pixels.
Total running time is about $10$ minutes per scene. } As an alternative to 3DGS, we also explored using NeRF for 3D representations, but empirically found that it inferior to 3DGS; see Sec.~\ref{subsec:nerflash}.


\begin{figure}[t!]
   \includegraphics[width=\columnwidth]{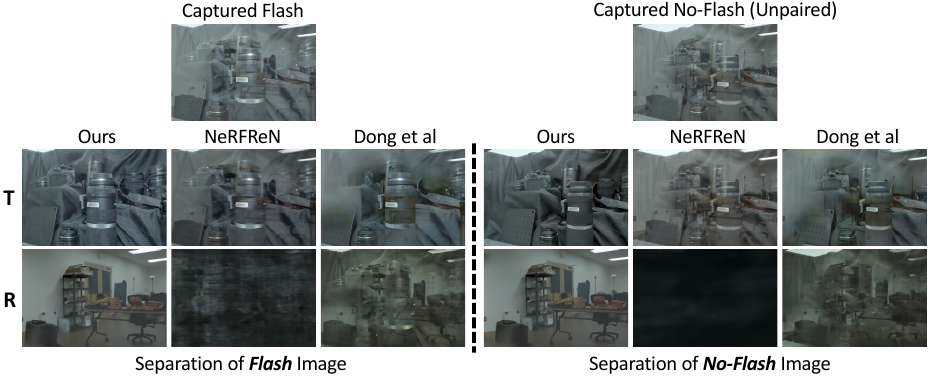}
   \caption{\textbf{The Lens Stage Scene.} Top, middle, and bottom rows are the captured images, separated transmissions, and separated reflections, respectively.  Our reflection separation approach is far more effective.}
   \label{fig:compare_baselines_lenstage}
\end{figure}

\begin{figure}[t!]
   \includegraphics[width=\columnwidth]{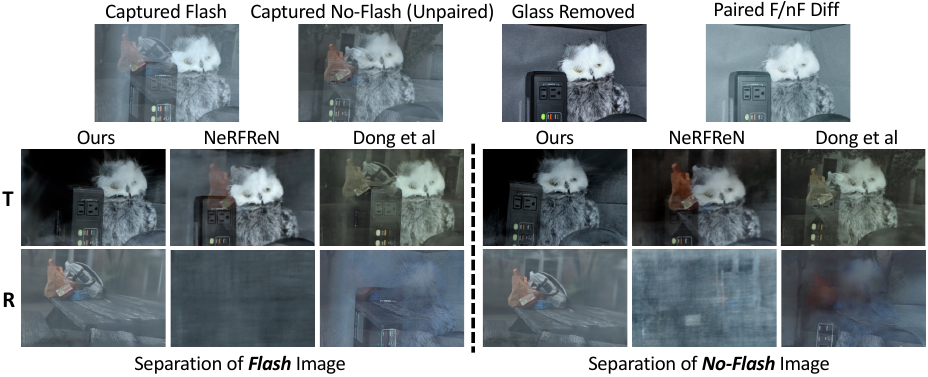}
   \caption{\textbf{The Outdoor Scene.} 
   The captured images, separated transmissions, and separated reflections are shown in the top, middle, and bottom rows, respectively.
   We also include two additional reference images in the top row as references. ``Paired F/nF Diff'' denotes the difference between flash and no-flash paired images. ``Glass Removed'' shows true transmission by directly removing the glass. 
   }
   \label{fig:compare_baselines_outdoor}
\end{figure}

\section{Results}

We compare our method's transmission-reflection separation ability against the baseline state-of-the-art methods. We also demonstrate novel view synthesis and depth estimation capabilities through our 3D inverse rendering framework.

\label{sec:result}
\topic{Transmission-Reflection Separation.}
\label{sec:visual_comparison}
Our method, Flash-Splat, outperforms both baselines in transmission-reflection separation on our real-world scenes. In fact, the baselines fail completely to produce a reasonable separation. Figures~\ref{fig:compare_baselines_office}, \ref{fig:compare_baselines_controller}, \ref{fig:compare_baselines_lenstage} show results on indoor scenes. \ref{fig:compare_baselines_outdoor} show results on an outdoor scene. Scene descriptions and results on 3 more scenes are included in the supplement. 



\begin{figure}[t!]
   \includegraphics[width=\columnwidth]{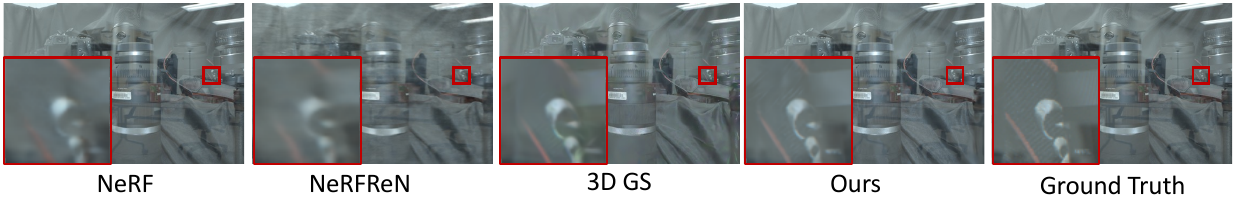}

   \caption{\textbf{Novel View Synthesis (NVS).} Our method does not compromise NVS quality, even when compared to dedicated NVS methods like NeRF and 3DGS. 
   }
   
   \label{fig:compare_baselines_novel_view}
\end{figure}

\topic{Novel View Synthesis (NVS).} Flash-Splat can perform novel view synthesis as it is based on inverse rendering. We compare our synthesis quality for scenes with reflections against NeRF~\cite{mildenhall2020nerf}, NeRFReN~\cite{guo2022nerfren}, and 3DGS (our implementation of 3DGS still follows FSGS~\cite{zhu2023fsgs}, as explained in Section~\ref{subsec:4.3_arch}). Figure~\ref{fig:compare_baselines_novel_view} shows Flash-Splat does not compromise NVS
performance, even when compared to dedicated NVS methods like NeRF and 3DGS.
Rendered videos of our separated transmitted and reflected 3D scenes are in our \href{https://flash-splat.github.io/}{\textcolor{magenta}{project webpage}}.

\begin{figure}[t!]
   \includegraphics[width=\columnwidth]{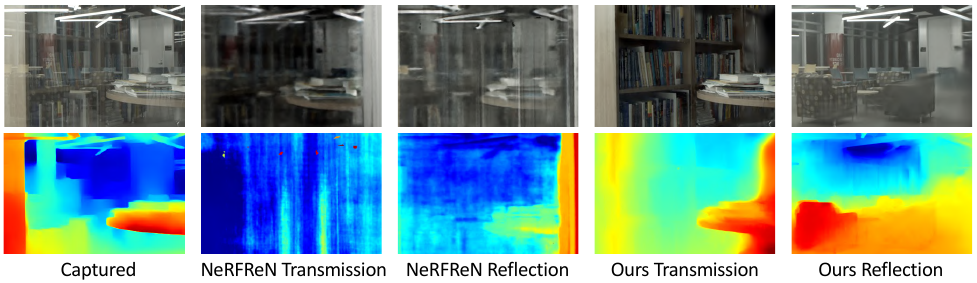}
   \caption{\textbf{Depth Estimation.} The captured image's depth estimated by MiDaS~\cite{Ranftl2020midas, Ranftl2021midas} is shown in the leftmost column, which cannot differentiate between transmitted and reflected scenes. Our depths are much better than NeRFReN's~\cite{guo2022nerfren}.
   }

   \label{fig:compare_baselines_depth}
\end{figure}

\topic{Depth Estimation.} Benefitting from the 3DGS base representation, Flash-Splat can also perform depth estimation on both transmitted and reflected scenes. Flash-Splat outperforms NeRFReN for both components (see Figure~\ref{fig:compare_baselines_depth}).





\begin{figure}[t!]
   \includegraphics[width=\columnwidth]{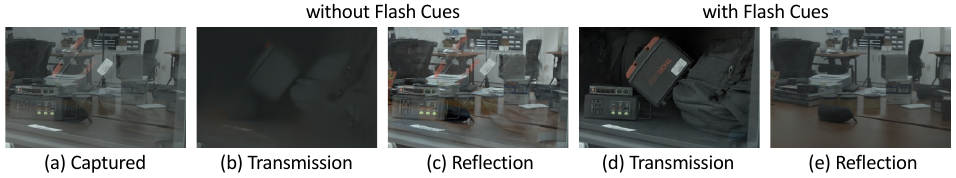}
   \caption{\textbf{With and Without the Flash Cues.} (\textcolor{blue}{b,c}) shows the reflection separation result if we do not utilize the flash cues in our proposed framework, details of which are discussed in Section~\ref{subsec:without_flash_Cues}. (\textcolor{blue}{d,e}) shows the result using the full version our proposed framework --- the separated reflection and transmission are significantly better. This highlights the importance of our proposed flash cues.
   }
   \label{fig:abla_cue_ours}
\end{figure}


\begin{figure}[t!]
   \includegraphics[width=\columnwidth]{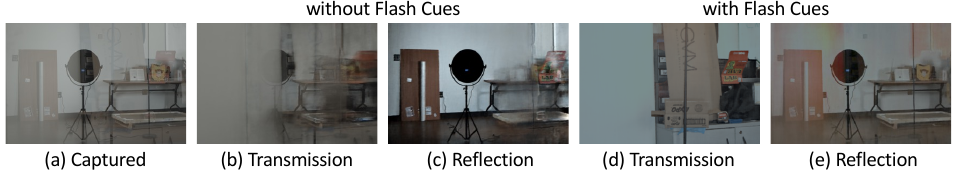}
   \caption{ \textbf{Equip NeRFReN~\cite{guo2022nerfren} With Flash Cues from 2D Pseudo-Pairs.} 
   (\textcolor{blue}{b,c}) show separation results from the original NeRFReN; (\textcolor{blue}{d,e}) show separation results from NeRFReN incorporated our proposed flash cues. Since NeRFReN is not compatible with the point clouds initialization using the 3D pseudo pair, we only equip NeRFReN with the Pearson linearity regularization from the 2D pseudo pairs. Our 2D pseudo pairs regularization alone can achieve better reflection separation performance than the original NeRFReN. 
   }
   \label{fig:abla_cue_nerfren}
\end{figure}



\section{Ablation Studies}
\label{sec:ablations}


\subsection{With and Without Flash Cues}
\label{subsec:without_flash_Cues}
To demonstrate the importance of flash cues, we design a ``flashless'' variant of our proposed framework where we remove the flash cues from the 2D and 3D pseudo-pairs. 
This ``flashless'' framework is still a 3DGS-based approach, but does not utilize flash/no-flash photography at all. Its detailed architecture is illustrated in {Figure \textcolor{blue}{19} in the supplement.} For a fair comparison, this flashless framework is using the same number of total views as our proposed framework. Figure~\ref{fig:abla_cue_ours} shows that this flashless framework achieves significantly worse separation performance than our proposed Flash-Splat framework, which highlights the indispensability of the flash cues.

\subsection{Replacing 3DGS with NeRF}
\label{subsec:nerflash}
To investigate the flash cues' impact on other 3D representations, we design another framework named Flash-NeRF, where we replace the 3DGS~\cite{kerbl20233dgs} representation with NeRF~\cite{mildenhall2020nerf} and keep everything else the same. Since NeRF is not compatible with our pseudo pair point clouds initialization, we only utilize the Pearson linearity regularization from the 2D pseudo pairs as the flash cue. The  Flash-NeRF framework can be seen as NeRFReN~\cite{guo2022nerfren} plus our 2D pseudo pairs regularization. Figure~\ref{fig:abla_cue_nerfren} shows that our 2D pseudo pairs regularization significantly enhances NeRFReN's reflection separation performance. 

Nevertheless, while this Flash-NeRF framework obtains an almost perfect transmitted scene, we can still notice obvious artifacts and floaters in the reflected scene. We find that using 3DGS as 3D representation can notably mitigate this issue, and thus decide to use 3DGS in our proposed framework.



\subsection{Discussions \& Limitations}
\topic{Reflections in COLMAP:} By Law of Reflection, reflected objects are equivalently ``virtual objects'' superimposed into the transmitted scene. We empirically verified reflections do not decrease COLMAP's view matching performance. \\
\topic{View Coverage:}~Although Flash-Splat does not need paired flash/no-flash images, it does require the flash and no-flash scenes to cover a similar range of perspectives. For instance, Flash-Splat would not work if we take flash images of the scene from the left and no-flash images from the right, as we would not be able to accurately synthesize pseudo-pairs in this case. \\
\topic{Curved Reflection:}~Flash-Splat currently cannot deal with scenes with curved reflective surfaces, e.g., curved glasses, since curved reflective surfaces will cause severe deformation of the reflected scene. We believe this is an important yet under-explored corner case for reflection removal, which we leave for future work.\\
\topic{Double Reflection:}~While double reflections can be a powerful cue for removing reflections when dealing with thick or double-pane glass~\cite{shih2015reflection}, Flash-Splat currently does not handle scenes with obvious double reflections. Incorporating double-reflection--based cues is an interesting research direction.\\
\topic{Flash Strength:}~If the flash is too weak to illuminate the transmission scene, Flash-Splat would not have the necessary flash cues to remove the reflection. \\
\topic{Dynamic Scene:}~Because our 3D representation is static, objects moving between captured images result in blurry reconstructions.

\section{Conclusion}
\label{sec:conclusions}

We present a novel approach for transmission-reflection separation of 3D scenes through flash cues, significantly mitigating the ill-posedness of the task. By synthesizing ``pseudo-paired'' flash/no-flash images within a 3D inverse rendering framework based on Gaussian Splatting, we demonstrate superior reflection separation capabilities, particularly under challenging conditions where traditional methods falter. We validate our method on a new real-world dataset, showcasing its effectiveness and robustness. Our method not only unlocks practical reflection-removal but also enables novel view synthesis and depth estimation separately for the transmitted and reflected 3D scene.

\subsection*{Acknowledgements}
This work was supported in part by AFOSR Young Investigator Program Award no.~FA9550-22-1-0208, ONR Award no.~N00014-23-1-2752, NSF CAREER Award no.~2339616, the Joint Directed Energy Transition Office, and a gift from Dolby Labs. We thank Kevin Zhang and Yi-Ting Chen for helpful discussions.

%
%
\bibliographystyle{splncs04}
\bibliography{main}
\newpage

\title{Flash-Splat: 3D Reflection Removal with \\ Flash Cues and Gaussian Splats -- Supplement} 
\setcounter{section}{0}
\setcounter{figure}{14}
\setcounter{table}{0}
\setcounter{equation}{9}
In this supplementary material, we show results on three additional scenes (Sec. \textcolor{blue}{1}), describe each scene's setup (Sec. \textcolor{blue}{2}), conduct more comparison experiments (Sec. \textcolor{blue}{3}), report quantitative performance (Sec. \textcolor{blue}{4}), and provide more details of an ablation study (Sec. \textcolor{blue}{5}). {Additionally, our \href{https://flash-splat.github.io/}{\textcolor{magenta}{project webpage}} shows the \textbf{rendered videos} of the transmitted and reflected 3D scenes separated by our proposed method, which outperforms the separation by NeRFReN~\cite{guo2022nerfren}. 


\section{Additional Scenes}
In addition to the 4 scenes we presented in Section~\textcolor{blue}{5}, we show results on three additional scenes. We evaluate our proposed method and the baselines on these scenes in the same way as described in Section~\textcolor{blue}{4}. As shown in Figure~\ref{fig:compare_baselines_shelf}, \ref{fig:compare_baselines_poster}, \ref{fig:compare_baselines_lab}, our proposed method significantly outperforms the baselines in terms of reflection transmission separation.

\section{Scene Descriptions}
In cases of strong specular reflections, such as the scenes in our captured dataset, it is challenging even for humans to identify which objects belong to the transmitted scene, and which belong to the reflected scene. Therefore, to help readers understand the setup of our scenes, we briefly describe the transmitted and reflected scenes for each of our 7 scenes (4 in the main paper, 3 in the supplement).

\begin{itemize}
    \item \textbf{Figure~\textcolor{blue}{5}, Office} (main paper). The transmitted scene is a bookcase in an office with a glass wall. We set up our camera in the corridor facing inside the office. The reflected scene is a study area at the end of the corridor.
    \item \textbf{Figure~\textcolor{blue}{6}, Game Controller} (main paper). The transmitted scene is a game controller in a black case with a glass cover. Note that the glass surface is horizontal to the ground. The reflected scene is a door with a glass window (you can also see the corridor through the door's window). The door is upside down due to reflection.
    \item \textbf{Figure~\textcolor{blue}{7}, Lens Stage} (main paper). The transmitted scene is a lens stage with several lenses on it. The lens stage is covered with a glass case. The reflected scene consists of tables and chairs. 
    \item \textbf{Figure~\textcolor{blue}{8}, Outdoor Scene} (main paper). We took photos of a toy and a power bank inside a glass window from outdoors. The reflected scene includes some bags on an outdoor table, with plants and another building's windows (mildly defocused) ~20 meters away in the background.
    \item \textbf{Figure~\ref{fig:compare_baselines_shelf}, Shelf} (supplement). The transmitted scene is a shelf with boxes, bags, and batteries on it. The reflected scene consists of tables and chairs.
    \item \textbf{Figure~\ref{fig:compare_baselines_poster}, Poster} (supplement). The transmitted scene is a poster on the wall of a corridor. The reflected scene is the corridor itself.
    \item \textbf{Figure~\ref{fig:compare_baselines_lab}, Lab} (supplement). The transmitted scene is a cabinet with some boxes on it and a lamp's pole (black) in front of it. The reflected scene includes a door, a lamp, and a table with various items on it.

\end{itemize}

\begin{figure}[h]
   \includegraphics[width=\columnwidth]{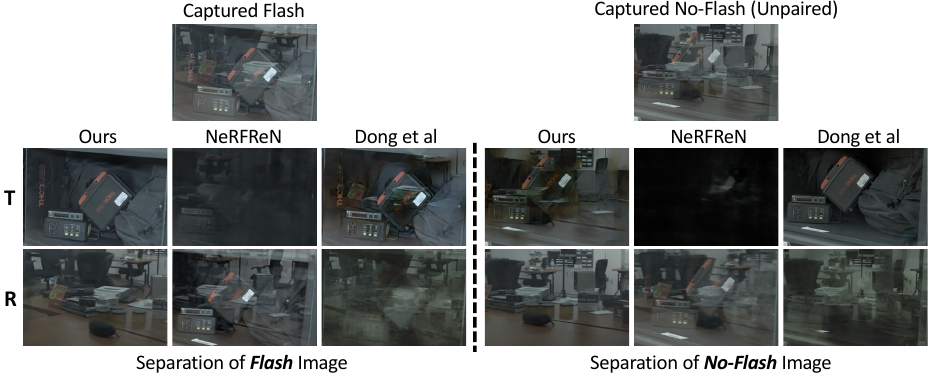}
   \caption{\textbf{Comparison with NeRFReN~\cite{guo2022nerfren} and Dong et al~\cite{dong2021location} on Shelf scene.} Top, middle, and bottom rows are the captured images, separated transmissions, and separated reflections, respectively. Our reflection separation approach is far more effective.}
   \label{fig:compare_baselines_shelf}
\end{figure}

\begin{figure}[]
   \includegraphics[width=\columnwidth]{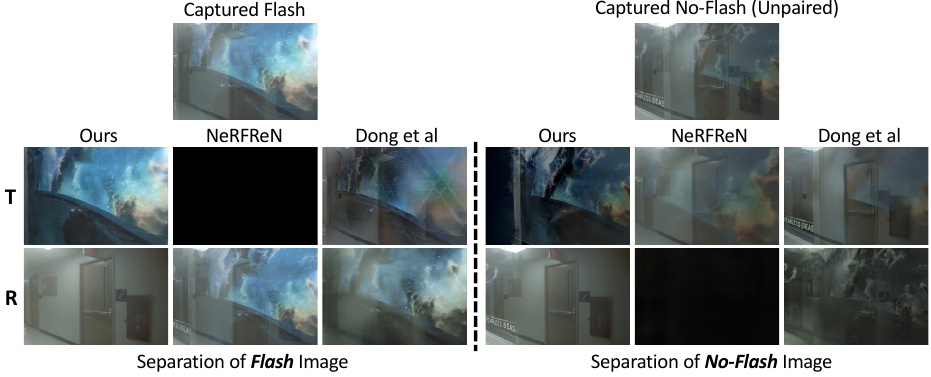}
   \caption{\textbf{Comparison with NeRFReN~\cite{guo2022nerfren} and Dong et al~\cite{dong2021location} on Poster scene.} Top, middle, and bottom rows are the captured images, separated transmissions, and separated reflections, respectively. Our reflection separation approach is far more effective.}
   \label{fig:compare_baselines_poster}
\end{figure}

\begin{figure}[]
   \includegraphics[width=\columnwidth]{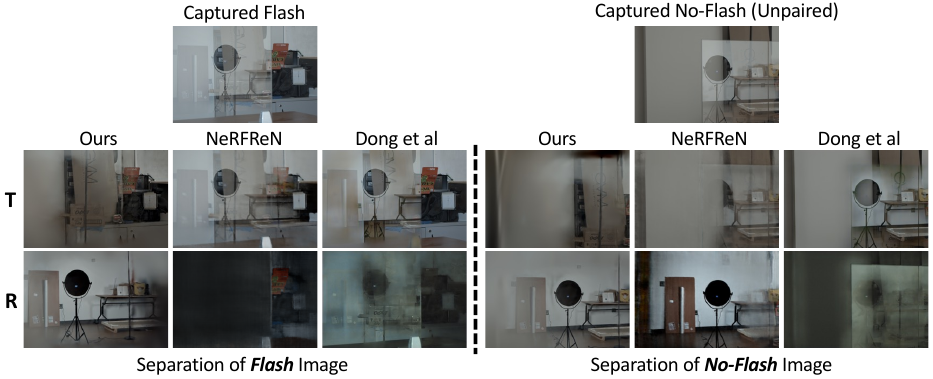}
   \caption{\textbf{Comparison with NeRFReN~\cite{guo2022nerfren} and Dong et al~\cite{dong2021location} on Lab scene.} Top, middle, and bottom rows are the captured images, separated transmissions, and separated reflections, respectively.  Our reflection separation approach is far more effective.}
   \label{fig:compare_baselines_lab}
\end{figure}


\begin{figure}[t!]
   \includegraphics[width=\columnwidth]{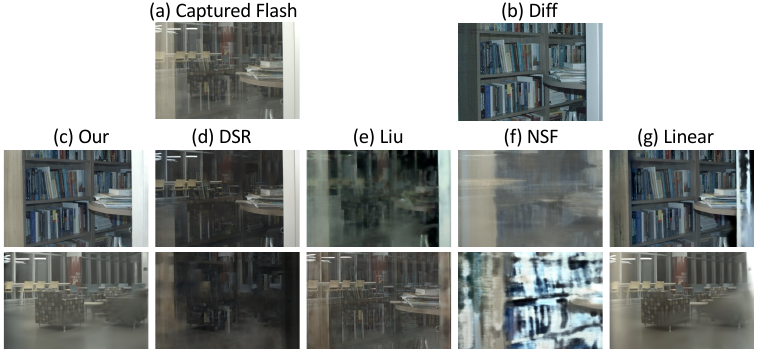}
   \caption{\textbf{Additional Comparisons.} Our method  (\textcolor{magenta}{c}) outperforms (\textcolor{magenta}{d}) DSRNet\cite{hu2023single} : ICCV 2023, supervised, single image-based; (\textcolor{magenta}{e}) Liu\cite{liu2020learning} : CVPR 2020, supervised, burst-based; and (\textcolor{magenta}{f}) Neural Spline Fields (NSF) \cite{chugunov2024nsf}: CVPR 2024, unsupervised, burst-based. Our reconstructed transmission is close to (\textcolor{magenta}{b}), the paired flash/no-flash difference (Diff), which requires paired images captured with a tripod. Additionally, in (\textcolor{magenta}{g}), we trained a pure linear representation by enforcing $T_F=c T_N$. This model results in imperfect reflection separation (notice the right side) compared to our soft-linear system with the Pearson loss. }
   \label{fig:more_baseline}
\end{figure}

\newpage
\section{Additional Comparisons}

We conduct 4 more comparison experiments.
As shown in Figure \ref{fig:more_baseline}, our method (\textcolor{magenta}{c}) also outperforms (\textcolor{magenta}{d}) DSRNet\cite{hu2023single} : ICCV 2023, supervised, single image-based; (\textcolor{magenta}{e}) Liu\cite{liu2020learning} : CVPR 2020, supervised, burst-based; and (\textcolor{magenta}{f}) Neural Spline Fields (NSF) \cite{chugunov2024nsf}: CVPR 2024, unsupervised, burst-based. Note that none of these methods can take advantage of our unpaired flash/no-flash data. Our reconstructed transmission is close to (\textcolor{magenta}{b}), the paired flash/no-flash difference (Diff), which requires paired images captured with a tripod. 

Additionally, as shown in Figure \ref{fig:more_baseline} (\textcolor{magenta}{g}), we trained a pure linear representation by enforcing $T_F=c T_N$. This model results in imperfect reflection separation (notice the right side) compared to our soft-linear system with the Pearson loss, since the relationship between $T_F$ and  $T_N$ is not perfectly linear.

\begin{table}[h!]
  \centering
  \footnotesize
  \renewcommand\tabcolsep{5pt}
  \begin{tabular}{@{}l|ccccc@{}}
    \toprule
    \multicolumn{1}{c}{}  & \multicolumn{5}{c}{Methods} \\
   
    Metric                    & \textbf{DSR}~\cite{hu2023single}        & \textbf{Liu}~\cite{liu2020learning}    & \textbf{NSF}~\cite{chugunov2024nsf} & \textbf{NeRFReN}~\cite{guo2022nerfren}  & \textbf{Ours}   \\
     \midrule
    PSNR  $\uparrow$          & 13.02                & 11.16           & 9.40         & 10.09             & \textbf{20.42}  \\
    LPIPS $\downarrow$        & 0.5754               & 0.6765          & 0.7452       & 0.7153            & \textbf{0.2868} \\
    \bottomrule
  \hline
  \end{tabular}
   \vspace{0.5cm}

  \caption{\textbf{Averaged Quantitative Evaluations.} We calculate the PSNR and LPIPS between each method's separated transmissions and the paired flash/no-flash differences, which serve as references for the ground truth transmissions. Our method has a huge quantitative advantage over the other methods, which corresponds with our huge qualitative advantage shown in the visual comparisons in Figure~\textcolor{blue}{5-8, 15-17}. Granted, our method's transmission is not perfect as it exhibits a slightly different color tone compared to the difference image, e.g., Figure~\ref{fig:more_baseline} (\textcolor{magenta}{b, c}). Nevertheless, our result successfully obtains structural information that is very close to the reference image, outperforming other methods by a large margin. }
  \label{tab:metrics}
\end{table}

\section{Quantitative Evaluation}
To compute quantitative metrics, we need to have a ground truth transmission scene as a reference. While it is difficult (oftentimes impossible) to remove the glass from a scene, we can instead compute the paired flash/no-flash difference as the reference transmission scene. In Table~\ref{tab:metrics}, we report the averaged PSNR and LPIPS between the difference image and each method's separated transmission scene. We find that our method performs the best.

\section{Details of the Ablation Study in Section~\textcolor{blue}{6.1}}
\label{sec:supp_details_of_ablation}
In Section~\textcolor{blue}{6.1} of the main paper, we design and test a flashless framework, where we remove the flash cues from our proposed framework and keep everything else the same. Figure~\ref{fig:pipeline_gs} shows the detailed architecture of this flashless framework.

\begin{figure}[h!]
   \includegraphics[width=\columnwidth]{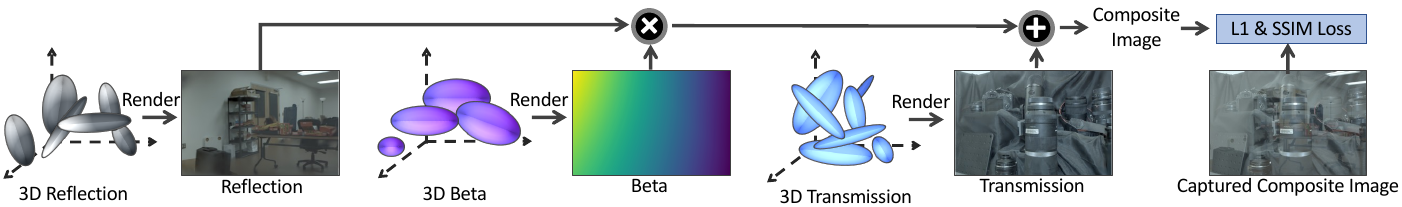}
   \caption{\textbf{Ablation: the flashless version of our proposed framework.} To demonstrate the importance of flash cues, we design an ablation study where we remove the flash cues from our proposed framework. This ``flashless'' framework is still a 3DGS-based approach, but does not utilize flash/no-flash photography at all. It uses 3 3DGSs to represent the reflected scene, the transmitted scene, and the reflection factor $\beta$. The loss is calculated between the captured images and images rendered from these 3 3DGSs. More descriptions can be found in Section~\textcolor{blue}{6.1} in the main paper.}
   \label{fig:pipeline_gs}
\end{figure}



\end{document}